\documentclass[final]{cvpr}

\usepackage{times}
\usepackage{epsfig}
\usepackage{graphicx}
\usepackage{amsmath}
\usepackage{amssymb}

\usepackage{subfigure}
\usepackage{csquotes}

\usepackage{multirow}
\usepackage{graphicx}
\usepackage{booktabs}
\usepackage{bbm}

\newcommand{\eqnsm}[2]{\begin{equation}\label{eq:#1}#2\end{equation}}

\usepackage{color, colortbl}

\newcommand{\TP}{\mathit{TP}}
\newcommand{\FN}{\mathit{FN}}
\newcommand{\FP}{\mathit{FP}}

\DeclareMathOperator*{\argmax}{argmax}

\newcommand*{\twoelementtable}[3][l]%
{%
\renewcommand{\arraystretch}{0.8}%
\begin{tabular}[t]{@{}#1@{}}%
#2\tabularnewline
#3%
\end{tabular}%
}
\usepackage{array}
\usepackage{tabularx}

\usepackage[pagebackref=true,breaklinks=true,colorlinks,bookmarks=false]{hyperref}

\def\cvprPaperID{3174} 
\def\confYear{CVPR 2021}

\begin{document}

\title{Cross-View Regularization for Domain Adaptive Panoptic Segmentation}

\author{Jiaxing Huang,\hspace{2mm} Dayan Guan,\hspace{2mm} Aoran Xiao,\hspace{2mm} Shijian Lu\thanks{Corresponding author (Shijian.Lu@ntu.edu.sg).} \vspace{0.5cm} \\ 
School of Computer Science Engineering, Nanyang Technological University\\
}

\maketitle

\begin{abstract}
Panoptic segmentation unifies semantic segmentation and instance segmentation which has been attracting increasing attention in recent years. However, most existing research was conducted under a supervised learning setup whereas unsupervised domain adaptive panoptic segmentation which is critical in different tasks and applications is largely neglected. We design a domain adaptive panoptic segmentation network that exploits inter-style consistency and inter-task regularization for optimal domain adaptive panoptic segmentation. The inter-style consistency leverages geometric invariance across the same image of the different styles which ` fabricates’ certain self-supervisions to guide the network to learn domain-invariant features. The inter-task regularization exploits the complementary nature of instance segmentation and semantic segmentation and uses it as a constraint for better feature alignment across domains. Extensive experiments over multiple domain adaptive panoptic segmentation tasks (e.g. synthetic-to-real and real-to-real) show that our proposed network achieves superior segmentation performance as compared with the state-of-the-art.
\end{abstract}

\section{Introduction}
\begin{figure}[t]
\centering
\subfigure {\includegraphics[width=0.98\linewidth]{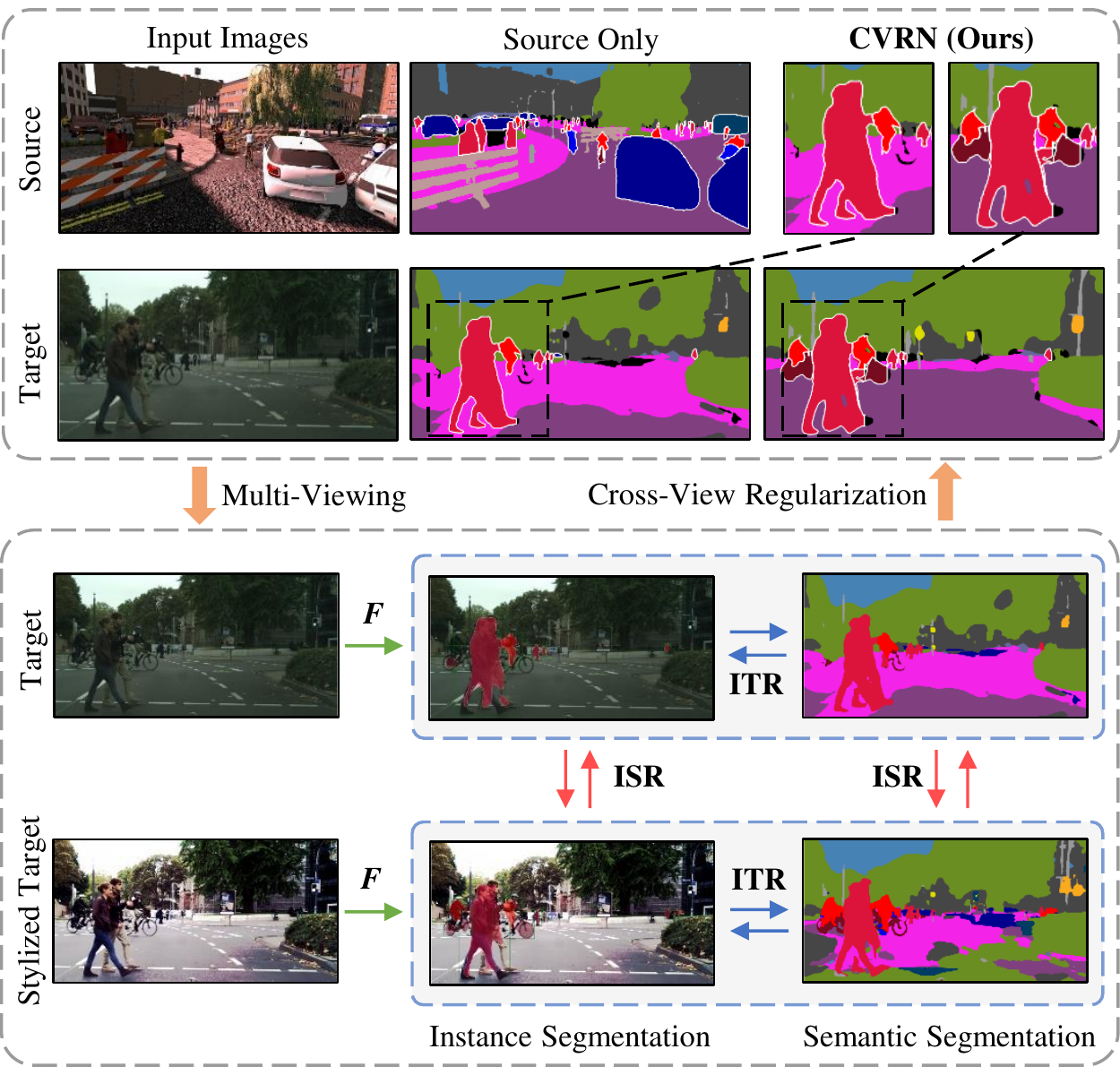}}
\caption{Our proposed cross-view regularization network (CVRN) tackles domain adaptive panoptic segmentation by exploring an inter-task regularization (ITR) and an inter-style regularization (ISR). As shown in the bottom part, ITR exploits the complementary nature of instance segmentation and semantic segmentation to regularize each other. ISR employs online image stylization to augments multiple views of the same image for regularization. Panoptic segmentations with and without our cross-view regularization are shown in the top part (the black dash lines highlight two close-up views). Best viewed in color.} 
\vspace{-10pt}
\label{fig:intro}
\end{figure}

Panoptic segmentation unifies semantic segmentation and instance segmentation which aims to assign a semantic class and an instance ID to each image pixel concurrently. With a large amount of annotated training images, panoptic segmentation has recently made rapid progress under a supervised setup~\cite{kirillov2019panoptic,kirillov2019PFPN,li2019attention,porzi2019seamless,liu2019end,xiong2019upsnet,Li_2020_CVPR,Hou_2020_CVPR,li2020unifying}. Unfortunately, collecting large-scale training images with pixel-level annotations is prohibitively expensive and time-consuming~\cite{cordts2016cityscapes,neuhold2017mapillary}. One way to mitigate this constraint is to leverage synthetic images \cite{ros2016synthia} that can be automatically annotated by graphic software. However, synthetic and natural images have clear domain gaps and panoptic segmentation models trained using synthetic images usually experience sharp performance drop while applied to natural images, as shown in Figure~\ref{fig:intro}.

One strategy that could better leverage synthetic images is domain adaptive panoptic segmentation that adopts certain unsupervised domain adaptation (UDA) techniques for adaptation from synthetic images to natural images. However, domain adaptive panoptic segmentation is largely neglected though domain adaptive semantic segmentation \cite{hong2018conditional, zhu2016beyond, hoffman2018cycada, sankaranarayanan2018learning, zou2018unsupervised, vu2019advent, zou2019confidence, Kundu_2019_ICCV, yang2020fda} and domain adaptive instance detection/segmentation \cite{chen2018domain,cai2019exploring,saito2019strong,vu2019advent,he2020domain,zou2019confidence,zhuang2020ifan,xu2020exploring,guan2021uncertainty}  have been investigated extensively. This could be due to a misconception that domain adaptive panoptic segmentation can be simply achieved by integrating domain adaptive semantic segmentation and domain adaptive instance segmentation. Nevertheless, semantic segmentation and instance segmentation are guided by different objectives which usually learn different feature representations from different perspectives. Learning the two tasks separately and then integrating the learned models is thus sub-optimal as it simply ignores the complementary nature of the two tasks.

We design CVRN, an innovative cross-view regularization network that addresses the challenge of domain adaptive panoptic segmentation through the regularization from different perspectives, as illustrated in Figure~\ref{fig:intro}. Instead of treating semantic segmentation and instance segmentation as two independent tasks in training, we designed an inter-task regularizer that guides the two tasks to complement and regularize each other to compensate the lack of annotations (for target-domain data) in domain adaptive panoptic segmentation. This design is inspired by our observations that semantic segmentation usually performs clearly better for amorphous regions called "stuff" as compared with countable objects called "things" whereas instance segmentation usually performs in an opposite manner. In addition, we designed an inter-style regularizer that formulates the geometry consistency of the same image across different styles (in illumination, weather conditions, contrast, etc.) as supervision to regularize domain adaptation and mitigate missing annotations in target domain. The inter-style regularizer treats different styles as different views of the same image which regularizes the domain adaptation within each single image. For both inter-task and inter-style regularization, we predict pseudo labels for target-domain samples by adapting self-training ideas that have been widely adopted in many other domain adaptive computer vision tasks~\cite{zou2018unsupervised, li2019bidirectional, zou2019confidence}.

The contributions of this work can be summarized in three aspects. \textit{First}, we designed a cross-view regularization network that addresses the challenge of domain adaptive panoptic segmentation effectively. To the best of our knowledge, this is the first work that tackles the challenging domain adaptive panoptic segmentation task. \textit{Second}, we designed a novel inter-task regularizer that exploits the complementary nature of semantic segmentation and instance segmentation for optimal domain adaptive panoptic segmentation. In addition, we designed an inter-style regularizer that formulates geometric consistency of the same image of different styles as supervision for better feature alignment across domains. \textit{Third}, extensive experiments over multiple domain adaptive panoptic segmentation tasks show that our network achieves superior segmentation performance as compared with the state-of-the-art. 

\begin{figure*}[ht]
\centering
\subfigure {\includegraphics[width=0.98\linewidth]{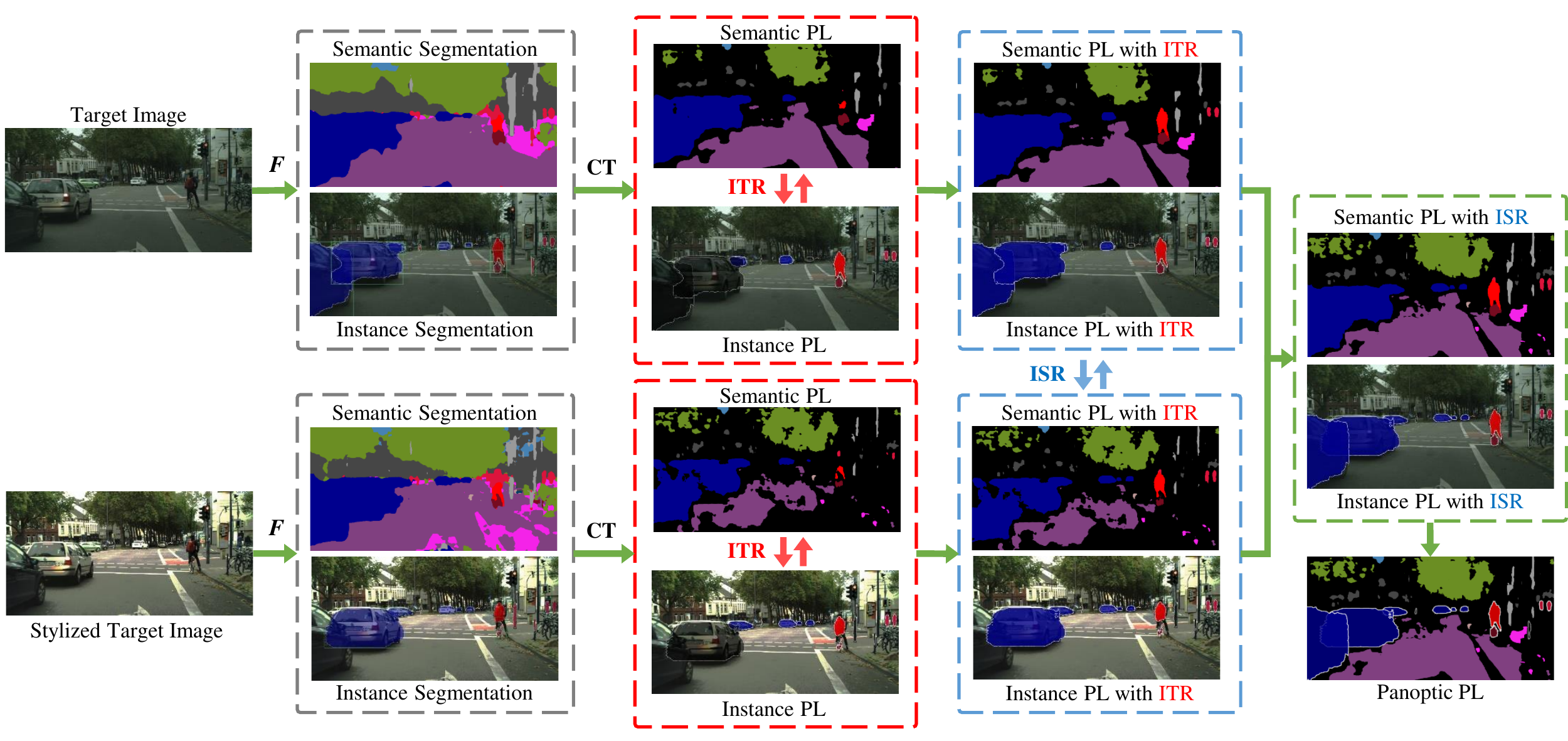}}
\caption{Overview of our proposed cross-view regularization network (CVRN): CVRN predicts multi-view co-regularized panoptic pseudo labels (PL) for learning from unlabelled target data. A target image and its stylized transformation are first fed to a panoptic segmentation model $F$ to generate predictions (in gray boxes) and four sets of primary pseudo labels (in \textcolor{red}{red} boxes). With the four sets of primary pseudo labels, inter-task regularization (ITR) exploits the complementary nature of instance segmentation and semantic segmentation to co-regularize their pseudo labels, e.g. the confident predictions in instance segmentation can guide the unconfident predictions in semantic segmentation (e.g., the rider), and vice versa. The inter-task regularized pseudo labels are shown in the \textcolor{blue}{blue} boxes. Similarly, inter-style regularization (ISR) exploits the complementary property of images of the same scene but different styles for regularization, where \textcolor{green}{green} box shows ISR regularized pseudo labels. Finally, the cross-view regularized instance and semantic segmentation pseudo labels are fused into panoptic segmentation pseudo labels to train unsupervised domain adaptive panoptic segmentation model with unlabeled target data.
} 
\label{fig:architecture}
\end{figure*}


\section{Related Works}

The concept of \textbf{Panoptic segmentation} was introduced in \cite{kirillov2019panoptic} which handles the problem by fusing the predictions of instance segmentation and semantic segmentation heuristically. Quite a number of relevant works have been reported since then. For example, \cite{kirillov2019PFPN} extends an instance segmentation model with a semantic segmentation branch and takes a shared feature pyramid network as backbone. \cite{li2019attention} employs instance-level attention to transfer knowledge from an instance segmentation branch to a semantic segmentation branch. \cite{liu2019end} presents a spatial ranking module to address occlusions between the predicted instances. \cite{xiong2019upsnet} introduces a non-parametric panoptic head for resolving the conflicts between instance and semantic segmentation. \cite{cheng2020panoptic} presents a bottom-up approach that employs a class-agnostic instance segmentation branch with center regression. 
Though panoptic segmentation has been studied extensively recently, most existing research was conducted under a supervised setup where all training data are fully annotated. We instead focus on a more challenging domain adaptive panoptic segmentation task that aims to adapt from an annotated source domain to an unlabelled target domain.

\textbf{Multi-view learning} trains a learner over two or more different views by incorporating confident predictions of target data iteratively \cite{blum1998combining,dasgupta2002pac, zhou2005tri,guan2019fusion,park2018adversarial}. In unsupervised domain adaptation, it generates pseudo labels for unlabelled target data for measuring and minimizing various task loss ($e.g.$, cross entropy loss in segmentation and detection) \cite{luo2008transfer, chen2011co,cao2019pedestrian,guan2019unsupervised}. In recent years, multi-view learning diversifies the learned parameters ($e.g.$, kernel weights) to enforce multiple classifiers via adversarial dropout \cite{saito2018adversarial,Lee_2019_ICCV}, classifier discrepancy maximization \cite{saito2018maximum,lee2019sliced}, parameter diversification \cite{zhang2018fully,luo2019taking}, asymmetric classifier tri-training \cite{saito2017asymmetric,he2020domain}, etc.
Different from existed multi-view learning that employs multiple classifiers to create multiple views in the feature space, we exploit the complementary nature of semantic segmentation and instance segmentation and use their predictions as two views in panoptic segmentation. Additionally, we adopt online image stylization to construct multiple views in the input space which enhances domain adaptation by enforcing geometry consistency across image styles.

\textbf{Unsupervised domain adaptation} aims to adapt a model from a labelled source domain to an unlabelled target domain which is very meaning for mitigating the data collection and annotation constraint in deep network training. One typical domain adaptation approach leverages adversarial learning that employs a domain classifier to learn domain invariant features \cite{ganin2014unsupervised,hoffman2016fcns,long2015learning,long2016unsupervised,tzeng2017adversarial,chen2018road,tsai2018learning,vu2019advent,guan2021scale,yang2020fda}. Another typical approach exploits self-training that predicts pseudo labels for target-domain data and includes confident predictions in network training \cite{zou2018unsupervised,saleh2018effective,zhu2005semi,gong2012geodesi,guan2019unsupervised,zhong2019invariance}. The self-training approach has attracted increasing attention in recent years, and different strategies have been designed for predicting high-quality pseudo labels by incorporating class-balance thresholding~\cite{zou2018unsupervised}, confidence-regularization~\cite{zou2019confidence}, voting-based densification~\cite{pan2020two} or scale-invariance example exploration~\cite{subhani2020learning}. 
We adopt the self-training idea in the implementation of our proposed cross-view regularization over different tasks and image styles. More specifically, we introduce multi-view learning (across tasks and image styles) into the self-training framework for unifying instance segmentation and semantic segmentation under the context of domain adaptive panoptic segmentation.

\section{Method}

This section presents the proposed cross-view regularization network (CVRN) as illustrated in Figure~\ref{fig:architecture}. Leveraging multi-task self-training (MTST), CVRN aims to adapt towards a domain-specific panoptic segmentation model by using unlabelled target images. It tackles cross-view regularization from two perspectives including inter-task regularization (ITR) and inter-style regularization (ISR), more details to be described in the ensuing subsections.

\subsection{Multi-Task Self-Training}
Our proposed cross-view regularization is implemented over a multi-task self-training (MTST) network. Self-training has been widely investigated to exploit unlabeled target-domain data in semi-supervised and unsupervised learning. It predicts pseudo labels for target-domain data and incorporates confident prediction for training stronger models iteratively. For the task of domain adaptive panoptic segmentation under study, we extend self-training to multi-task self-training and use it as a base to implement our proposed inter-task regularization between semantic segmentation and instance segmentation. 

The problem setting is as follows. We have source-domain images $X_{s} \subset \mathbb{R}^{H \times W \times 3}$ and the corresponding pixel-level semantic labels $Y_{s} \subset (1,C,N)^{H \times W}$ where $C$ denotes the class number and $N$ denotes the ID index of things {\ie, countable objects}. We also have unlabelled  target-domain images $X_{t} \subset \mathbb{R}^{H \times W \times 3}$. The target is to learn a panoptic segmentation model $F$ that performs well in the unlabeled target domain. With $X_{s}$ and $Y_{s}$ in the source domain, $F$ can be optimized by a supervised panoptic segmentation loss $\mathcal{L}_{pan}$ that consists of a semantic segmentation loss $\mathcal{L}_{seg}$ and an instance segmentation loss $\mathcal{L}_{ins}$. During training, $Y_{s} \subset (1,C,N)^{H \times W}$ are converted to semantic segmentation labels $Y^{e}_{s} \subset (1,C)^{H \times W}$ and instance segmentation labels $Y^{d}_{s} \subset (1,C^{th},N)^{H \times W}$ for computing $\mathcal{L}_{seg}$ and $\mathcal{L}_{ins}$ concurrently. 

For $x_{t}$ in the target domain, $F$ will generate panoptic segmentation predictions which are a heuristic combination of semantic segmentation predictions $p^{e}_{t}$ and instance segmentation predictions $p^{d}_{t}$. 
Based on generated predictions $p^{e}_{t}$ and $p^{d}_{t}$, pseudo labels can be determined by a selection function $\mathcal{S}$ (confidence thresholding) that is defined as follows:
\begin{equation}
\begin{split}
\mathcal{S}({{p_{t}}}) = \argmax_{c \in C} \mathbbm{1}_{[p_{t}^{(c)} > \exp(-k_{c})]}(p_{t}^{(c)})
\end{split}
\end{equation}
where $p_{t}$ refers to either $p^{e}_{t}$ or $p^{d}_{t}$, $\mathbbm{1}$ is a function that returns the input if the condition is true or an empty output otherwise, and $k_{c}$ is the class-balanced weights~\cite{zou2018unsupervised}. The panoptic segmentation model $F$ can be retained with target-domain images $X_{t}$ and the predicted pseudo labels $\mathcal{S}(P^{e}_{t})$ and $\mathcal{S}(P^{d}_{t})$ via self-training.

\textbf{Supervised loss:}  
Given the panoptic segmentation model $F$, a source-domain image $x_{s} \subset X_{s}$ and its corresponding semantic segmentation and instance segmentation labels $\{y^{e}_{s},y^{d}_{s}\} \subset \{Y^{e}_{s}, Y^{d}_{s}\}$, the supervised panoptic segmentation loss $\mathcal{L}_{pan}$ can be defined as follows:
\begin{equation}
\begin{split}
\mathcal{L}_{pan}(x_{s},y_{s};F) = 
 \mathcal{L}_{seg}(p^{e}_{s},y^{e}_{s}) +  \mathcal{L}_{ins}(p^{d}_{s},y^{d}_{s}),
\end{split}
\end{equation}
where $\mathcal{L}_{seg}$ is cross-entropy loss, $\mathcal{L}_{ins}$ is instance segmentation loss as defined in \cite{he2017mask}, $p^{e}_{s}$ and $p^{d}_{s}$ represent source-domain semantic segmentation and instance segmentation predictions, respectively.

\textbf{Multi-task self-training loss:}  
Given the panoptic segmentation model $F$ and a target image $x_{t} \subset X_{t}$, the multi-task self-training loss $\mathcal{L}_{mtst}$ can be defined by:
\begin{equation}
\begin{split}
\mathcal{L}_{mtst}(x_{t};F) = 
 \mathcal{L}_{seg}(p^{e}_{t},\mathcal{S}({p^{e}_{t}})) + 
 \mathcal{L}_{ins}(p^{d}_{t},\mathcal{S}({p^{d}_{t}})),
\end{split}
\end{equation}
where $p^{e}_{t}$ and $p^{d}_{t}$ represent target-domain semantic segmentation and instance segmentation predictions, respectively.

\subsection{Inter-Task Regularization}
Semantic segmentation and instance segmentation learn from different perspectives which often produce complementary predictions under the context of panoptic segmentation. Under the framework of multi-task self-training, we observe that pseudo labels predicted by the two tasks often complement each other. As illustrated in Figure~\ref{fig:architecture}, semantic segmentation tends to predict high-quality pseudo labels (\ie, diverse and accurate) for amorphous regions called ``stuff" but low-quality pseudo labels (\ie, sparse and inaccurate) for countable objects called ``things". On the contrary, instance segmentation tends to generate high-/low-quality pseudo labels for things/stuff. 

Based on this observation, we design an inter-task regularization (ITR) method that mutually regularize self-training in between the semantic segmentation and instance segmentation tasks in the target domain. Specifically, ITR employs high-certainty (\ie, low entropy) pseudo label predictions of one task to regularize pseudo label predictions of the other task and vice versa. In this way, ITR is capable of predicting higher quality pseudo labels as compared with those predicted by each single task alone. 

In the unannotated target domain, pseudo labels of semantic segmentation and instance segmentation (\ie, $\mathcal{S}({p^{e}_{t}})$ and $\mathcal{S}({p^{d}_{t}})$) can be determined based on the predictions (\ie, $p^{e}_{t}$ and $p^{d}_{t}$) as described in the last subsection. $\mathcal{S}({p^{d}_{t}})$ can then be regularized by $\mathcal{S}({p^{e}_{t}})$. This instance segmentation pseudo-label regularization function $\mathcal{R}_{d}$ that is defined as follows:
\begin{equation}
\begin{split}
\mathcal{R}_{d}({p^{d}_{t}},{p^{e}_{t}}) = 
& \mathbbm{1}_{[\mathcal{E}({p^{d}_{t}}) < \mathcal{E}({p^{e}_{t}})]}(\mathcal{S}({p^{d}_{t}})) + \\
& \argmax_{c \in C} \mathbbm{1}_{[\mathcal{J}(p^{d}_{t},\mathcal{S}({p^{e}_{t}}))]}({p^{d(c)}_{t}}),
\end{split}
\end{equation}
where $\mathcal{J}$ is a function to judge if the instance segmentation prediction $p^{d}_{t}$ is highly consistent with the semantic segmentation pseudo label $\mathcal{S}({p^{e}_{t}})$ in the same image location.
Similarly, $\mathcal{S}({p^{e}_{t}})$ can be regularized by $\mathcal{S}({p^{d}_{t}})$. This semantic segmentation pseudo-label regularization function $\mathcal{R}_{e}$ is defined by:
\begin{equation}
\begin{split}
\mathcal{R}_{e}(p^{e}_{t},{p^{d}_{t}}) = 
& \mathbbm{1}_{[\mathcal{E}({p^{e}_{t}}) < \mathcal{E}({p^{d}_{t}})]}(\mathcal{S}({p^{e}_{t}})) + \\
& \mathcal{T} ( \mathbbm{1}_{[\mathcal{E}({p^{d}_{t}}) < \mathcal{E}({p^{e}_{t}})]}(\mathcal{S}({p^{d}_{t}}))),
\end{split}
\end{equation}
where $\mathcal{E}$ is the entropy function as defined in~\cite{shannon1948mathematical}.
$\mathcal{T}$ is a label transformation function from instance segmentation to semantic segmentation,
\ie, ignoring ID indexes of instances of the same category.

\textbf{Inter-task regularization loss:}
Given the panoptic segmentation model $F$ and a target-domain image $x_{t} \subset X_{t}$, the inter-task regularization loss $\mathcal{L}_{itr}$ can be defined by:
\begin{equation}
\begin{split}
\mathcal{L}_{itr}(x_{t};F) = 
& \mathcal{L}_{seg}(p^{e}_{t},\mathcal{R}_{e}(p^{e}_{t},{p^{d}_{t}})) + \\
& \mathcal{L}_{ins}(p^{d}_{t},\mathcal{R}_{d}((p^{d}_{t},{p^{e}_{t}})),
\end{split}
\end{equation}
where $p^{e}_{t}$ and $p^{d}_{t}$ represent target-domain semantic segmentation and instance segmentation predictions, respectively.

\subsection{Inter-style regularization}
Besides inter-task regularization, we also design an inter-style regularization (ISR) method that further improves the quality of pseudo labels by fusing confident (pseudo label) predictions of images of the same scene but different styles. The idea is that images of the same scene should share perfectly the same pixel-level semantics when they are captured under different conditions with different image styles (e.g. in different illumination, weather, etc.). A pixel with more confident pseudo-level prediction in one image view with one specific style can therefore be exploited to regularize the less confident prediction of the corresponding pixel in another image view with a different style.

We implement ISR based on online image stylization that generates images with new styles/views with histogram matching \cite{pizer1987adaptive}. Specifically, the online image stylization first transfers a training image $x_{t}$ to $\widetilde{x_{t}}$ based on another randomly selected target-domain image.  It then forwards $x_{t}$ and its style transformation $\widetilde{x_{t}}$ to the panoptic segmentation model $F$ to predict semantic segmentation (\ie, $p^{e}_{t}$ and $\widetilde{p}^{e}_{t}$) and instance segmentation (\ie, $p^{d}_{t}$ and $\widetilde{p}^{d}_{t}$). Finally, we generate the pseudo labels by an inter-style pseudo-label unification function that is defined as follows:
\begin{equation}
\begin{split}
\mathcal{U}(p_{t}, \widetilde{p_{t}}) =
& \mathbbm{1}_{[\mathcal{E}(p_{t}) < \mathcal{E}(\widetilde{p_{t}})]}(\mathcal{S}(p_{t})) + \\
& \mathbbm{1}_{[\mathcal{E}(\widetilde{p_{t}}) < \mathcal{E}(p_{t}))]}(\mathcal{S}(\widetilde{p_{t}})) .
\end{split}
\end{equation}
ISR thus enforces geometry consistency across images of the same contents but different styles by retraining the model $F$ with unified pseudo labels in different style views. 

\textbf{Inter-style regularization loss:}
Given the panoptic segmentation model $F$ and a target-domain image $x_{t} \subset X_{t}$ and its style transformation $\widetilde{x_{t}}$, the inter-style regularization loss $\mathcal{L}_{isr}$ can be formulated as follows:
\begin{equation}
\begin{split}
\mathcal{L}_{isr}(x_{t} ,\widetilde{x_{t}}; F) = 
& \mathcal{L}_{seg}(\widetilde{p^{e}_{t}},\mathcal{U}(p^{e}_{t}, \widetilde{p^{e}_{t}})) + \\
& \mathcal{L}_{ins}(\tilde{p}^{d}_{t},\mathcal{U}(p^{d}_{t}, \widetilde{p^{d}_{t}})),
\end{split}
\end{equation}
where $\tilde{p}^{e}_{t}$ and $\tilde{p}^{d}_{t}$ represent semantic segmentation and instance segmentation predictions generated from style-transformed images, respectively.

\textbf{Training objective:}
The overall objective function of the proposed cross-view regularization network (CVRN) can thus be formulated by summing up the four training losses as follows:
\begin{equation}
\begin{split}
\mathcal{L}_{cvrn} =  \mathcal{L}_{pan} + \lambda_{mt}  \mathcal{L}_{mtst} + \lambda_{it} \mathcal{L}_{itr} +  \lambda_{is} \mathcal{L}_{isr},
\end{split}
\label{eq:cvrn}
\end{equation}
where $\lambda_{mt}$, $\lambda_{it}$ and $\lambda_{is}$ is the balancing weights.

\section{Experiment}

\subsection{Datasets and Evaluation Metrics}
We evaluated our proposed cross-view regularization technique over three widely used datasets: 

\textbf{SYNTHIA \cite{ros2016synthia}} is a large-scale synthetic dataset with 9,400 images that are generated by random perturbation of virtual environments. This dataset provides pixel-level annotations for semantic segmentation as well as object-level labels for instance segmentation. Panoptic segmentation annotations can be obtained by fusing ``stuff" regions as annotated for semantic segmentation with object labels as annotated for instance segmentation. All the images have the same resolution of $760 \times 1280$.

\textbf{Cityscapes \cite{cordts2016cityscapes}} is a widely used autonomous driving dataset with images captured by an image acquisition system mounted in a driving vehicle. It consists of 2,975 training images and 500 validation images with dense manual annotations for panoptic segmentation. All the images have the same resolution of $1024 \times 2048$.

\textbf{Mapillary Vistas \cite{neuhold2017mapillary}} is a large-scale autonomous driving dataset with images captured by different image acquisition sensors. It consists of 18,000 training images and 2,000 validation images with high-quality annotations for panoptic segmentation. The resolution of the dataset image varies from $768\times1024$ to $4000 \times 6000$.

We evaluate panoptic segmentation by three widely used metrics including semantic quality (SQ), recognition quality (RQ), and panoptic quality (PQ)~\cite{kirillov2019panoptic}. For each object class, PQ is actually the product of SQ and RQ:
 \eqnsm{psq-seg-det}{\small{\text{PQ}} = \underbrace{\frac{\sum_{(p, y) \in \TP} \text{IoU}(p, y)}{\vphantom{\frac{1}{2}}|\TP|}}_{\text{segmentation quality (SQ)}} \times \underbrace{\frac{|\TP|}{|\TP| + \frac{1}{2} |\FP| + \frac{1}{2} |\FN|}}_{\text{recognition quality (RQ) }} ,}
where $p$ is predicted segmentation and $y$ is ground truth. $\TP$, $\FP$ and $\FN$ denote true positives, false positives, and false negatives, which define matched segmentation (IoU 	$>$ 0.5~\cite{kirillov2019panoptic}), unmatched predicted segmentation (IoU $\leq$ 0.5 with any ground truth), and unmatched ground truth segmentation (IoU $\leq$ 0.5 with any predictions), respectively.

\renewcommand\arraystretch{1.2}
\begin{table}[t]
\centering
\begin{footnotesize}
\begin{tabular}{p{1.8cm}|*{4}{p{0.5cm}}|*{3}{p{0.3cm}}}
\toprule
 \multicolumn{8}{c}{\textbf{SYNTHIA $\rightarrow$ Cityscapes}} \\
 \midrule
Method &$L_{sup}$ &$L_{mtst}$ &$L_{itr}$ &$L_{isr}$ &mSQ &mRQ & PQ \\
\midrule
Source only &\checkmark &  &  &  &58.9 &29.7 &21.8 \\
MTST &\checkmark &\checkmark  &  &  &59.5 &32.9 &24.8  \\ 
MTST + ITR &\checkmark &\checkmark  &\checkmark &  &61.1  &36.4  &28.0  \\
MTST + ISR &\checkmark &\checkmark  & &\checkmark  &62.3  &36.9  &28.9  \\ 
\textbf{CVRN} &\checkmark &\checkmark &\checkmark  &\checkmark  &\textbf{66.6} &\textbf{40.9} &\textbf{32.1}  \\
\bottomrule
\end{tabular}
\end{footnotesize}
\caption{Ablation study of CVRN over domain adaptive panoptic segmentation task SYNTHIA $\rightarrow$ Cityscapes: The proposed inter-task regularization (ITR) and inter-style regularization (ISR) both outperforms the base network MTST (multi-task self-training) greatly. ITR and ISR complement with each other clearly.}
\label{tab:abla}
\end{table}

\renewcommand\arraystretch{1.2}
\begin{table*}[t]
\centering
\begin{footnotesize}
\begin{tabular}{p{2cm}|*{16}{p{0.3cm}}*{3}{p{0.4cm}}}
 \toprule
 \multicolumn{20}{c}{\textbf{SYNTHIA $\rightarrow$ Cityscapes Panoptic Segmentation}} \\
 \midrule
 Methods  &{road} &{side.} &{buil.}&{wall} &{fence} &{pole} &{light} &{sign} &{vege.} &{sky} &{pers.} &{rider} &{car} &{bus} &{mot.} &{bike} &mSQ  &mRQ &mPQ\\
 \midrule
 Source only &32.3 &5.1 &58.5 &0.9 &0.0 &0.9 &0.0 &4.6 &61.7 &61.3 &27.6 &9.5 &32.8 &22.6 &1.0 &2.7 &59.0 &27.8 &20.1 \\
 FDA~\cite{yang2020fda} &79.0 &22.0 &61.8 &1.1 &0.0 &5.6 &5.5 &9.5 &51.6 &70.7 &23.4 &16.3 &\textbf{34.1} &31.0 &5.2 &8.8 &65.0 &35.5 &26.6 \\
 CRST~\cite{zou2019confidence} &75.4 &19.0 &70.8 &1.4 &0.0 &7.3 &0.0 &5.2 &74.1 &69.2 &23.7 &\textbf{19.9} &33.4 &26.6 &2.4 &4.8 &60.3 &35.6 &27.1 \\
 AdvEnt~\cite{vu2019advent}  &\textbf{87.1} &32.4 &69.7 &1.1 &0.0 &3.8 &0.7 &2.3 &71.7 &72.0 &\textbf{28.2} &17.7 &31.0 &21.1 &6.3 &4.9 &65.6 &36.3 &28.1 \\
\textbf{CVRN (Ours)} &86.6 &\textbf{33.8} &\textbf{74.6} &\textbf{3.4} &0.0 &\textbf{10.0} &\textbf{5.7} &\textbf{13.5} &\textbf{80.3} &\textbf{76.3} &26.0 &18.0 &\textbf{34.1} &\textbf{37.4} &\textbf{7.3} &\textbf{6.2} &\textbf{66.6} &\textbf{40.9} &\textbf{32.1} \\
 \midrule
 \midrule
 \multicolumn{20}{c}{\textbf{SYNTHIA $\rightarrow$ Mapillary Panoptic Segmentation}} \\
 \midrule
 Methods  &{road} &{side.} &{buil.}&{wall} &{fence} &{pole} &{light} &{sign} &{vege.} &{sky} &{pers.} &{rider} &{car} &{bus} &{mot.} &{bike} &mSQ  &mRQ &mPQ\\
 \midrule
 Source only &22.1 &5.5 &34.5 &0.2 &0.0 &2.8 &0.0 &1.5 &40.3 &\textbf{79.5} &18.9 &9.2 &35.6 &3.9 &0.9 &0.5 &59.4 &21.3 &16.0 \\
 AdvEnt~\cite{vu2019advent} &27.7 &6.1 &28.1 &0.3 &0.0 &3.4 &1.6 &5.2 &48.1 &66.5 &28.4 &13.4 &40.5 &14.6 &5.2 &3.3 &63.6 &24.7 &18.3 \\
 CRST~\cite{zou2019confidence} &36.0 &6.4 &29.1 &0.2 &0.0 &2.8 &0.5 &4.6 &47.7 &68.9 &28.3 &13.0 &42.4 &13.6 &5.1 &2.0 &63.9 &25.2 &18.8 \\
 FDA~\cite{yang2020fda} &\textbf{44.1} &7.1 &26.6 &1.3 &0.0 &3.2 &0.2 &5.5 &45.2 &61.3 &30.1 &13.9 &39.4 &12.1 &\textbf{8.5} &7.0 &63.8 &26.1 &19.1 \\
\textbf{CVRN (Ours)} &33.4 &\textbf{7.4} &\textbf{32.9} &\textbf{1.6} &0.0 &\textbf{4.3} &0.4 &\textbf{6.5} &\textbf{50.8} &76.8 &\textbf{30.6} &\textbf{15.2} &\textbf{44.8} &\textbf{18.8} &7.9 &\textbf{9.5} &\textbf{65.3} &\textbf{28.1} &\textbf{21.3} \\
 \midrule
 \midrule
 \multicolumn{20}{c}{\textbf{Cityscapes $\rightarrow$ Mapillary Panoptic Segmentation}} \\
 \midrule
 Methods  &{road} &{side.} &{buil.}&{wall} &{fence} &{pole} &{light} &{sign} &{vege.} &{sky} &{pers.} &{rider} &{car} &{bus} &{mot.} &{bike} &mSQ  &mRQ &mPQ\\
 \midrule
 Source only &75.6 &15.8 &40.4 &5.1 &8.9 &3.2 &2.1 &13.4 &55.2 &81.8 &24.4 &15.1 &51.4 &4.4 &12.8 &13.6 &71.2 &34.3 &26.4 \\
  CRST~\cite{zou2019confidence} &77.0 &22.6 &40.2 &7.8 &10.5 &5.5 &11.3 &21.8 &56.5 &77.6 &29.4 &18.4 &56.0 &27.7 &11.9 &18.4 &72.4 &39.9 &30.8 \\
 FDA~\cite{yang2020fda} &74.3 &\textbf{23.4} &42.3 &9.6 &11.2 &6.4 &\textbf{15.4} &23.5 &60.4 &78.5 &33.9 &19.9 &52.9 &8.4 &\textbf{17.5} &16.0 &72.3 &40.3 &30.9 \\
 AdvEnt~\cite{vu2019advent}  &76.2 &20.5 &42.6 &6.8 &9.4 &4.6 &12.7 &24.1 &59.9 &83.1 &34.1 &\textbf{22.9} &54.1 &16.0 &13.5 &18.6 &72.7 &40.3 &31.2 \\
\textbf{CVRN (Ours)} &\textbf{77.3} &21.0 &\textbf{47.8} &\textbf{10.5} &\textbf{13.4} &\textbf{7.5} &14.1 &\textbf{25.1} &\textbf{62.1} &\textbf{86.4} &\textbf{37.7} &20.4 &\textbf{55.0} &\textbf{21.7} &14.3 &\textbf{21.4} &\textbf{73.8} &\textbf{42.8} &\textbf{33.5} \\
\bottomrule
\end{tabular}
\end{footnotesize}
\caption{Comparing CVRN with state-of-the-art domain adaptive panoptic segmentation: CVRN outperforms the state-of-the-art across three tasks. PQ is computed for each category. Mean SQ (mSQ), mean RQ (mSQ), mean PQ (mPQ) are computed over all categories.}
\label{tab:bench1}
\end{table*}

\subsection{Implementation Details} We adopted PSN~\cite{kirillov2019panoptic} as the panoptic segmentation architecture that consists of a semantic segmentation branch (\ie, \textit{Deeplab-V2}~\cite{chen2017deeplab}) and a instance segmentation branch (\ie, \textit{Mask R-CNN}~\cite{he2017mask}).All the networks in the experiments use ResNet-101~\cite{he2016deep} pre-trained on ImageNet~\cite{deng2009imagenet} as backbone. We implemented CVRN by using PyTorch~\cite{paszke2017automatic} and trained it with a single NVIDIA 2080TI GPU with 11GB memory. The networks are trained with a standard Stochastic Gradient Descent optimizer~\cite{bottou2010large} with learning rate $2.5\times 10^{-4}$, momentum $0.9$, and weight decay $10^{-4}$. The balancing weights $\lambda_{mt}$, $\lambda_{it}$ and $\lambda_{is}$ are both set as 1 in all experiments.

\subsection{Ablation studies}
\label{sec:abla}
We performed extensive ablation studies to investigate how our designs contribute to domain adaptive panoptic segmentation. The ablation studies were conducted over the task SYNTHIA $\rightarrow$ Cityscapes as shown in Table~\ref{tab:abla}. Specifically, we trained 5 models including: 1) \textit{Source only} that uses the supervised loss $L_{pan}(x_s, y_s; F)$ with no adaptation; 2) \textit{MTST} that uses the multi-task self-training loss $L_{mtst}(x_t; F)$ and $L_{pan}(x_s, y_s; F)$; 3) \textit{MTST+ITR} that uses the inter-task regularization loss $L_{itr}(x_t; F)$ , $L_{mtst}(x_t; F)$ and $L_{pan}(x_s, y_s; F)$; 4) \textit{MTST+ISR} that uses the inter-style regularization loss $L_{isr}(x_t; F)$, $L_{mtst}(x_t; F)$ and $L_{pan}(x_s, y_s; F)$; and 5) \textit{CVRN} that uses all four losses $L_{isr}(x_t; F)$, $L_{itr}(x_t; F)$, $L_{mtst}(x_t; F)$ and $L_{pan}(x_s, y_s; F)$.

Table~\ref{tab:abla} shows experimental results. It can be seen that \textit{MTST} outperforms \textit{Source only} clearly as it exploits target-domain knowledge via self-training. In addition, \textit{MTST+ITR} outperforms \textit{MTST} by a clear margin, demonstrating the effectiveness of the proposed inter-task regularization in between instance segmentation and semantic segmentation. At the same time, \textit{MTST+ISR} outperforms \textit{MTST} significantly as well which is largely attributed to the proposed inter-style regularization design that aggregates rich and complementary knowledge from different image views. Further, \textit{CVRN} produces the best panoptic segmentation, which shows that the proposed inter-task regularization and inter-style regularization are orthogonal and complementary to each other.

\renewcommand\arraystretch{1.2}
\begin{table*}[t]
\centering
\begin{footnotesize}
\begin{tabular}{p{2cm}|*{16}{p{0.3cm}}p{0.5cm}p{0.6cm}}
 \toprule
 \multicolumn{19}{c}{\textbf{SYNTHIA $\rightarrow$ Cityscapes Semantic Segmentation}} \\
 \midrule
 Methods  &{road} &{side.} &{buil.}&{wall} &{fence} &{pole} &{light} &{sign} &{vege.} &{sky} &{pers.} &{rider} &{car} &{bus} &{mot.} &{bike} &mIoU  &mIoU*\\
 \midrule
 Source only &56.8 &26.2 &68.6 &6.3 &0.4 &23.9 &3.6 &11.9 &76.5 &75.7 &37.4 &14.4 &50.5 &12.7 &8.2 &33.0 &31.6 &36.6\\
 PatAlign~\cite{tsai2019domain} &82.4 &38.0 &78.6 &8.7 &0.6 &26.0 &3.9 &11.1 &75.5 &84.6 &53.5 &21.6 &71.4 &32.6 &19.3 &31.7 &40.0 &46.5\\
 AdaptSeg~\cite{tsai2018learning} &84.3 &42.7 &77.5 &- &- &- &4.7 &7.0 &77.9 &82.5 &54.3 &21.0 &72.3 &32.2 &18.9 &32.3 &- &46.7\\
 CLAN~\cite{luo2019taking} &81.3 &37.0 &{80.1} &- &- &- &{16.1} &{13.7} &78.2 &81.5 &53.4 &21.2 &73.0 &32.9 &{22.6} &30.7 &- &47.8\\
 AdvEnt~\cite{vu2019advent} &85.6 &42.2 &79.7 &{8.7} &0.4 &25.9 &5.4 &8.1 &{80.4} &84.1 &{57.9} &23.8 &73.3 &36.4 &14.2 &{33.0} &41.2 &48.0\\
 IDA~\cite{pan2020unsupervised} &84.3 &37.7 &79.5 &5.3 &0.4 &24.9 &9.2 &8.4 &80.0 &84.1 &57.2 &23.0 &78.0 &38.1 &20.3 &36.5 &41.7 &48.9\\
 TIR~\cite{kim2020learning} &\textbf{92.6} &\textbf{53.2} &79.2 &- &- &- &1.6 &7.5 &78.6 &84.4 &52.6 &20.0 &82.1 &34.8 &14.6 &39.4 &- &49.3\\ 
 CrCDA~\cite{huang2020contextual}  &{86.2}	&{44.9}	&79.5	&8.3	&{0.7}	&{27.8}	&9.4	&11.8	&78.6	&\textbf{86.5}	&57.2	&{26.1}	&{76.8}	&{39.9}	&21.5	&32.1	&{42.9}	&{50.0}\\
 CRST~\cite{zou2019confidence} &67.7 &32.2 &73.9 &10.7 &\textbf{1.6} &\textbf{37.4} &22.2 &\textbf{31.2} &80.8 &80.5 &60.8 &29.1 &\textbf{82.8} &25.0 &19.4 &45.3 &43.8 &50.1\\
 BDL~\cite{li2019bidirectional} &86.0   &46.7   &80.3&-&-&-&14.1   &11.6 &79.2 &81.3 &54.1   &27.9   &73.7   &\textbf{42.2}   &25.7   &45.3  &- &51.4\\
 SIM~\cite{wang2020differential} &83.0 &44.0 &80.3 &- &- &- &17.1 &15.8 &80.5 &81.8 &59.9 &\textbf{33.1} &70.2 &37.3 &28.5 &45.8 &- &52.1\\
 FDA~\cite{yang2020fda} &79.3 &35.0 &73.2 &9.1 &0.3 &33.5 &19.9 &24.0 &61.7 &82.6 &\textbf{61.4} &31.1 &83.9 &40.8 &\textbf{38.4} &51.1 &45.3 &52.5\\
\textbf{CVRN (Ours)} &87.5 &45.5 &\textbf{83.5} &\textbf{12.2} &0.5 &\textbf{37.4} &\textbf{25.1} &29.6 &\textbf{85.9} &86.0 &61.1 &25.9 &80.9 &34.7 &33.8 &\textbf{53.5} &\textbf{49.0} &\textbf{56.6}
\\ 
\bottomrule
\end{tabular}
\end{footnotesize}
\caption{Comparisons of CVRN with state-of-the-art domain adaptive semantic segmentation: CVRN outperforms the state-of-the-art by large margins. IoU is evaluated for each of 16 pixel categories, \textit{mIoU*} is evaluated for 13 pixel classes following \cite{tsai2018learning,luo2019taking,li2019bidirectional,kim2020learning,wang2020differential}.}
\label{tab:bench2}
\end{table*}

\subsection{Comparison with state-of-art}
Since there is little domain adaptive panoptic segmentation work, we compare our method with a number of domain adaptation methods \cite{vu2019advent,zou2019confidence,yang2020fda} that achieved  state-of-the-art performance in both semantic segmentation and instance detection/segmentation. These methods can be easily applied to the panoptic segmentation task by heuristically combining predictions from semantic and instance segmentation~\cite{kirillov2019panoptic}
The comparisons were conducted over three domain adaptive panoptic segmentation tasks as shown in Table~\ref{tab:bench1}. We can observe that CVRN achieves the best panoptic segmentation across all three evaluation metrics for the task \enquote{SYNTHIA $\rightarrow$ Cityscapes} (synthetic-to-real). The superior panoptic segmentation is largely attributed to the proposed cross-view regularization that guides to exploit more confident samples which leads to more true positives. For another two tasks \enquote{SYNTHIA $\rightarrow$ Mapillary Vistas} (synthetic-to-real) and \enquote{SYNTHIA $\rightarrow$ Mapillary Vistas} (real-to-real), CVRN outperforms the state-of-the-art consistently due to similar reasons.

We also compare CVRN with state-of-the-art domain adaptive semantic segmentation methods \cite{tsai2018learning,tsai2019domain,luo2019taking,vu2019advent,pan2020unsupervised,kim2020learning,huang2020contextual,zou2019confidence,li2019bidirectional,wang2020differential,yang2020fda} and domain adaptive instance detection/segmentation methods \cite{chen2018domain,vu2019advent,saito2019strong,zou2019confidence,yang2020fda} that are dedicated to the two specific tasks, respectively. We compare with the two categories of methods because there is little domain adaptive panoptic segmentation work but panoptic segmentation actually consists of semantic segmentation and instance detection/segmentation. Tables ~\ref{tab:bench2} and \ref{tab:bench3} show experimetal results over the task \enquote{SYNTHIA $\rightarrow$ Cityscapes}. As Table~\ref{tab:bench2} shows, CVRN outperforms state-of-the-art domain adaptive semantic segmentation methods by a large margin (over 3.7 in mIoU). For domain adaptive instance detection and segmentation task, CVRN outperforms the state-of-the-art with a mAP of 2.7 for instance detection and a mAP of 5.4 for instance segmentation, respectively, as shown in Table \ref{tab:bench3}.

\begin{table}[t]
\renewcommand{\arraystretch}{1.2}
\centering
\begin{footnotesize}
\begin{tabular}{p{2cm}|*{6}{p{0.4cm}}|p{0.6cm}}
 \toprule
  \multicolumn{8}{c}{\textbf{SYNTHIA $\rightarrow$ Cityscapes Instance Detection}} \\
 \midrule
 Methods &{pers.} &{rider} &{car} &{bus} &{mot.} &{bike} &mAP\\
 \midrule
 Source only &42.9 &23.4 &38.4 &20.4 &3.5 &17.1 &24.3  \\
 DA \cite{chen2018domain} &43.2 &44.0 &39.9 &22.8 &7.6 &23.6 &30.2  \\
 SWDA \cite{saito2019strong} &{43.9} &35.4 &42.1 &28.5 &11.3 &26.4 &31.3  \\
 AdvEnt \cite{vu2019advent} &{43.9} &39.6 &44.0 &22.2 &11.2 &26.8 &31.3  \\
 CRST \cite{zou2019confidence} &40.3 &\textbf{48.6} &34.8 &29.7 &10.7 &28.6 &32.1  \\
 FDA \cite{yang2020fda} &42.6 &43.9 &42.5 &24.9 &10.6 &\textbf{30.1} &32.4  \\
 CRDA \cite{xu2020exploring} &\textbf{45.5} &37.9 &\textbf{45.6} &28.2 &9.1 &29.1 &32.6  \\
 \textbf{CVRN (Ours)} &43.5 &41.9 &\textbf{45.2} &{39.3} &\textbf{14.8} &27.1 &\textbf{35.3}  \\
 \midrule
 \midrule
  \multicolumn{8}{c}{\textbf{SYNTHIA $\rightarrow$ Cityscapes Instance Segmentation}} \\
 \midrule
 Methods &{pers.} &{rider} &{car} &{bus} &{mot.} &{bike} &mAP\\
 \midrule
 Source only &30.4 &9.2 &32.5 &20.4 &3.1 &1.3 &16.2  \\
 DA \cite{chen2018domain} &30.4 &\textbf{25.9} &28.4 &13.6 &4.3 &1.0 &17.3  \\
 SWDA \cite{saito2019strong} &30.3 &16.8 &31.8 &25.6 &4.1 &0.7 &18.2 \\
 AdvEnt \cite{vu2019advent} &30.8 &19.9 &32.0 &19.6 &7.0 &4.2 &18.9  \\
 FDA \cite{yang2020fda} &27.7 &24.6 &33.9 &22.5 &5.5 &5.4 &19.9 \\
CRDA \cite{xu2020exploring} &33.9 &16.7 &34.3 &27.7 &3.8 &3.0 &19.9  \\
 CRST \cite{zou2019confidence} &26.4 &20.3 &31.5 &27.9 &8.4 &8.6 &20.5  \\
 \textbf{CVRN (Ours)} &\textbf{34.4} &25.3 &\textbf{38.7} &\textbf{38.1} &\textbf{10.1} &\textbf{8.7} &\textbf{25.9}  \\
\bottomrule
\end{tabular}
\end{footnotesize}
\caption{Quantitative comparison of CVRN with state-of-the-art domain adaptive instance detection and instance segmentation methods: Class-wise AP (\%) and mAP (\%) over all classes are computed with an IoU threshold of 0.5 following \cite{chen2018domain,saito2019strong,xu2020exploring}.}
\label{tab:bench3}
\end{table}

\begin{figure*}[ht]
\centering
\begin{minipage}[h]{0.245\linewidth}
\centering\small {Ground Truth}
\end{minipage}
\vspace{2pt}
\begin{minipage}[h]{0.245\linewidth}
\centering\small {Source only} 
\end{minipage}
\begin{minipage}[h]{0.245\linewidth}
\centering\small {Advent}
\end{minipage}
\begin{minipage}[h]{0.245\linewidth}
\centering\small {\textbf{CVRN (Ours)}} 
\end{minipage}
\vspace{2pt}
\centering
\begin{minipage}[h]{0.245\linewidth}
\centering\includegraphics[width=.99\linewidth]{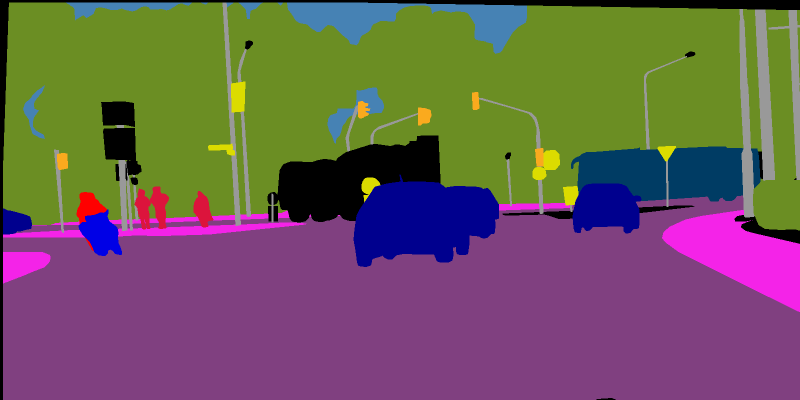}
\end{minipage}
\begin{minipage}[h]{0.245\linewidth}
\centering\includegraphics[width=.99\linewidth]{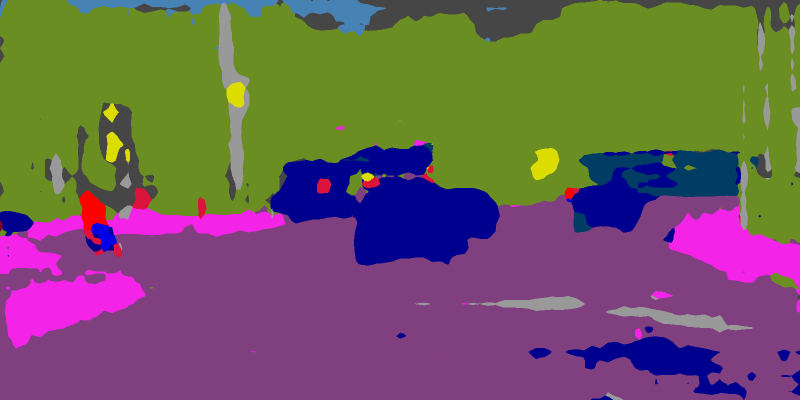}
\end{minipage}
\begin{minipage}[h]{0.245\linewidth}
\centering\includegraphics[width=.99\linewidth]{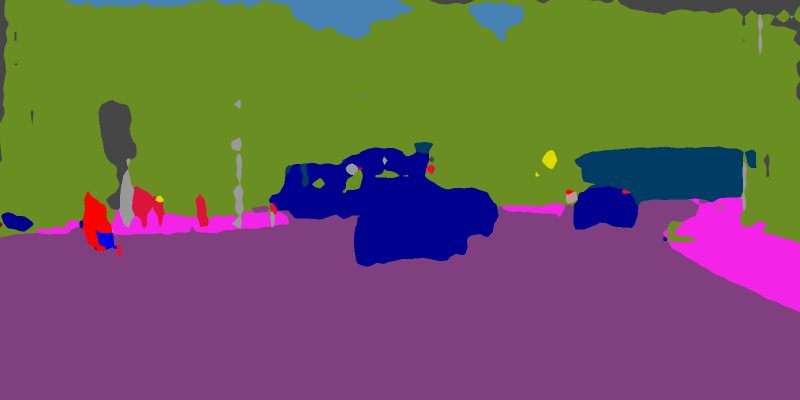}
\end{minipage}
\begin{minipage}[h]{0.245\linewidth}
\centering\includegraphics[width=.99\linewidth]{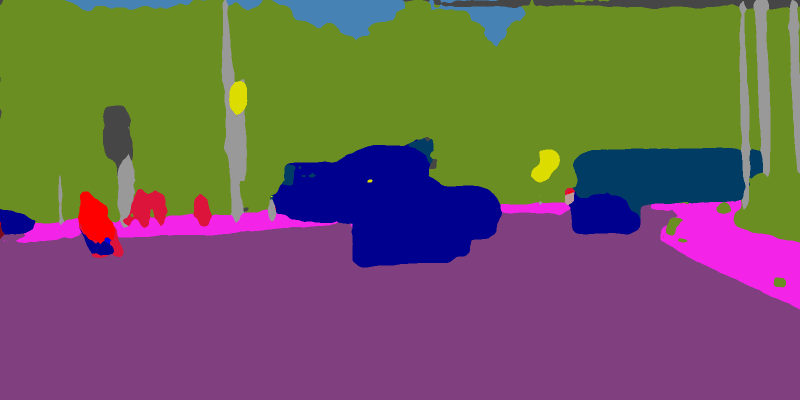}
\end{minipage}
\centering
\vspace{2pt}
\centering
\begin{minipage}[h]{0.245\linewidth}
\centering\includegraphics[width=.99\linewidth]{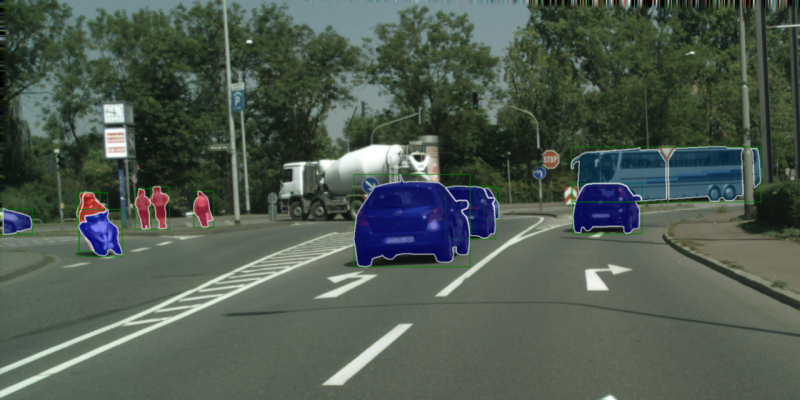}
\end{minipage}
\begin{minipage}[h]{0.245\linewidth}
\centering\includegraphics[width=.99\linewidth]{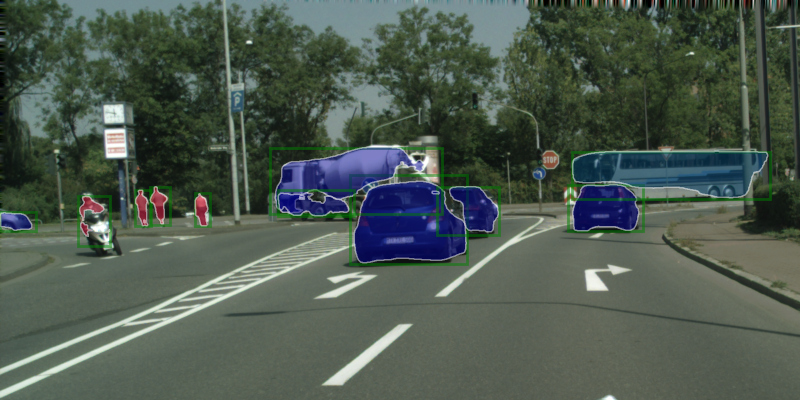}
\end{minipage}
\begin{minipage}[h]{0.245\linewidth}
\centering\includegraphics[width=.99\linewidth]{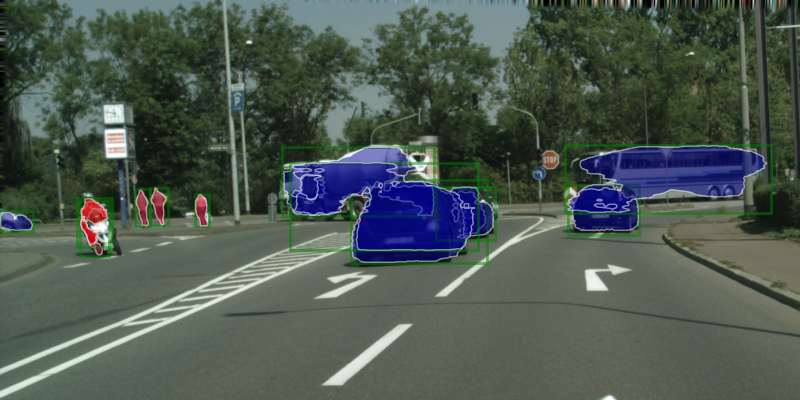}
\end{minipage}
\begin{minipage}[h]{0.245\linewidth}
\centering\includegraphics[width=.99\linewidth]{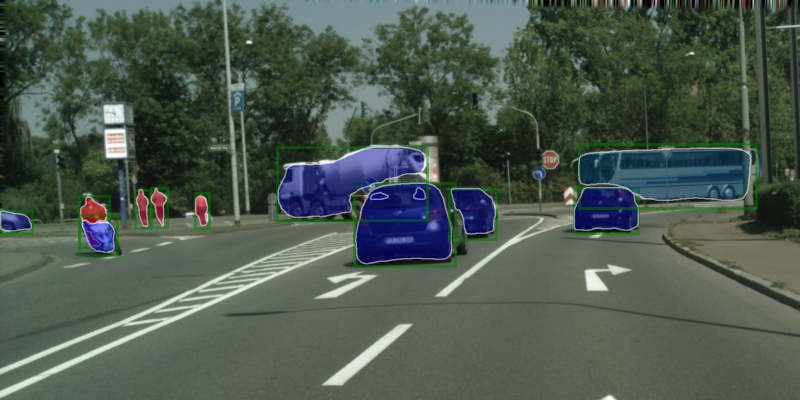}
\end{minipage}
\centering
\vspace{2pt}
\centering
\begin{minipage}[h]{0.245\linewidth}
\centering\includegraphics[width=.99\linewidth]{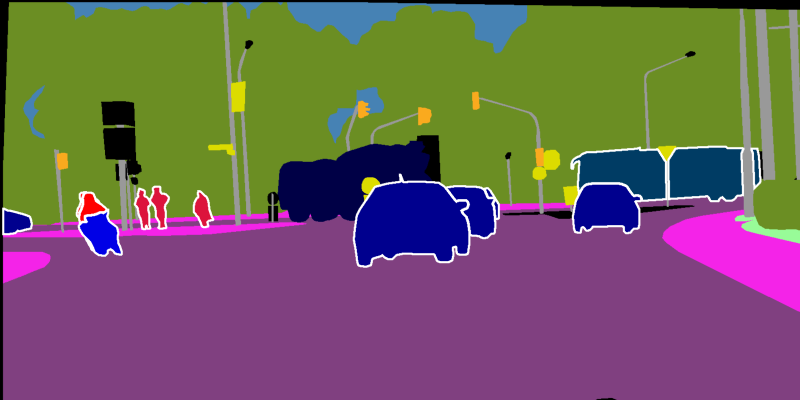}
\end{minipage}
\begin{minipage}[h]{0.245\linewidth}
\centering\includegraphics[width=.99\linewidth]{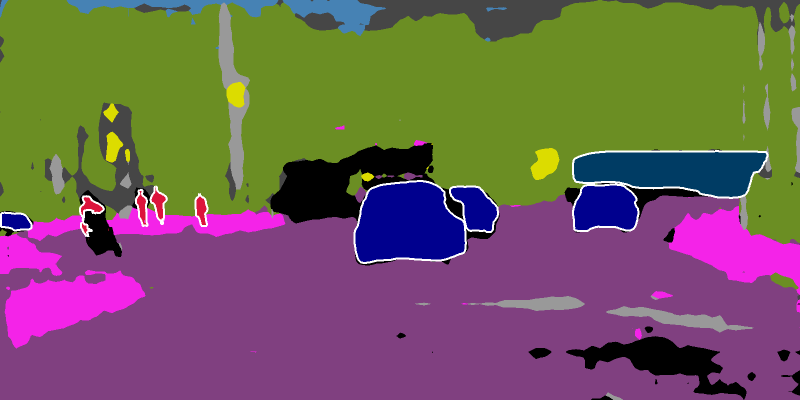}
\end{minipage}
\begin{minipage}[h]{0.245\linewidth}
\centering\includegraphics[width=.99\linewidth]{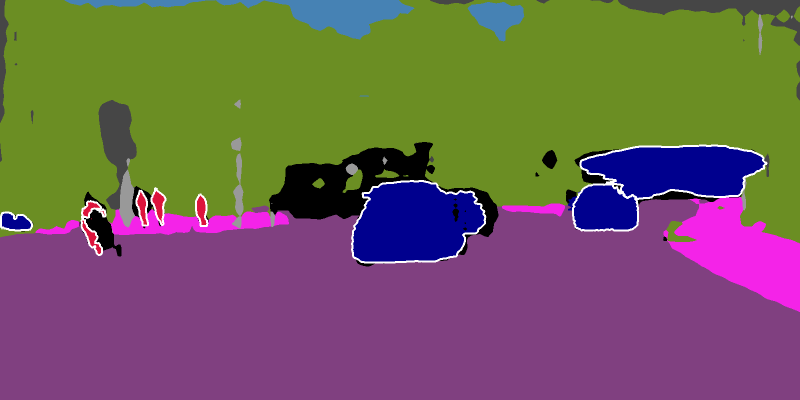}
\end{minipage}
\begin{minipage}[h]{0.245\linewidth}
\centering\includegraphics[width=.99\linewidth]{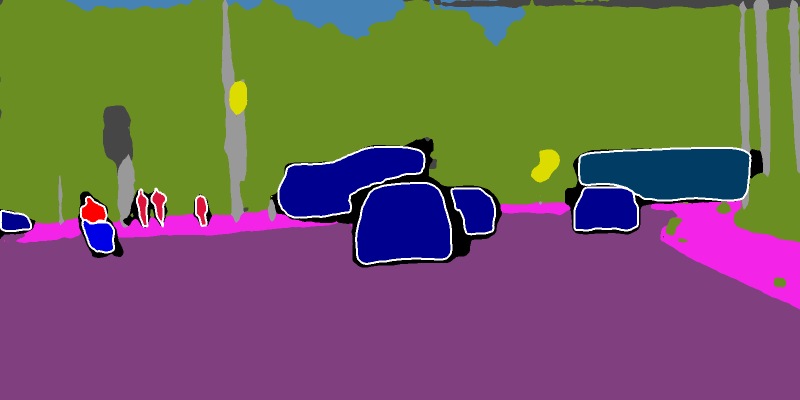}
\end{minipage}
\centering
\vspace{2pt}
\caption{Qualitative comparison of CVRN with \textit{Source only} (with no adaptation) and state-of-the-art method \textit{AdvEnt} \cite{vu2019advent} for domain adaptive semantic segmentation, instance segmentation and panoptic segmentation as shown in Rows 1-3, respectively. The comparison was conducted over the task \enquote{SYNTHIA $\rightarrow$ Cityscapes}. Best viewed in color.}
\vspace{-10pt}
\label{fig:result}
\end{figure*}

The qualitative segmentation is well aligned with the quantitative experimental results as illustrated in Figure~\ref{fig:result}. We compared CVRN with the baseline \textit{Source only} and state-of-the-art method \textit{Advent} \cite{vu2019advent} qualitatively over the domain adaptation task \enquote{SYNTHIA $\rightarrow$ Cityscapes}. As Figure~\ref{fig:result} shows, CVRN identifies and detects more correct ``thing" instances and segments more accurate ``stuff" regions across semantic segmentation, instance detection/segmentation and panoptic segmentation consistently. The superior segmentation performance is largely attributed to the proposed cross-view regularization that guides to learn rich and complementary semantic information from multiple views.

\subsection{Discussion}

We studied whether the proposed CVRN is complementary with state-of-the-art UDA methods for the task of domain adaptive panoptic segmentation. In the experiments, we incorporated our designed cross-view regularizers into each studied domain adaptation method and train panoptic segmentation models over the task SYNTHIA $\rightarrow$ Cityscapes. Table \ref{tab:comp} shows experimental results. We can observe that the incorporation of CVRN improves segmentation consistently across all studied UDA methods~\cite{vu2019advent,zou2019confidence,yang2020fda} and evaluation metrics. For metric mPQ that is more relevant to panoptic segmentation, the performance gains are above 3.3 for all three UDA methods. Such experimental results show that CVRN is complementary to existing UDA methods that work by image translation (\eg, FDA~\cite{yang2020fda}), adversarial learning (\eg, AdvEnt~\cite{vu2019advent}) and single-view self-training (\eg, CRST~\cite{zou2019confidence}).

\renewcommand\arraystretch{1.2}
\begin{table}[t]
\centering
\begin{footnotesize}
\begin{tabular}{p{1.25cm}|*{2}{p{0.25cm}}p{0.45cm}|*{2}{p{0.25cm}}p{0.45cm}|*{2}{p{0.25cm}}p{0.45cm}}
\toprule
 \multicolumn{10}{c}{\textbf{SYNTHIA $\rightarrow$ Cityscapes Panoptic Segmentation}} \\
 \midrule
\multicolumn{1}{c|}{Method} &\multicolumn{3}{c|}{Base}  &\multicolumn{3}{c|}{+ CVRN}  &\multicolumn{3}{c}{Gain} \\
&mSQ &mRQ &mPQ &mSQ &mRQ &mPQ &mSQ &mRQ &mPQ \\
\midrule
FDA~\cite{yang2020fda} &65.0 &35.5 &26.6  &66.0  &39.5  &30.9  &+1.0  &+4.0  &+3.3  \\
AdvEnt~\cite{vu2019advent} &65.6 &36.3 &28.1 &66.6  &40.5  &32.0  &+1.0  &+4.2  &+3.9  \\
CRST~\cite{zou2019confidence} &60.3 &35.6 &27.1  &66.6 &41.5 &32.4  &+6.3  &+5.9  &+5.3 \\
\bottomrule
\end{tabular}
\end{footnotesize}
\vspace{2pt}
\caption{CVRN complements with existing UDA methods: \textit{CVRN} can be easily incorporated into state-of-the-art representative UDA methods~\cite{vu2019advent,zou2019confidence,yang2020fda} with consistent performance improvement (evaluated over domain adaptive panoptic segmentation task SYNTHIA $\rightarrow$ Cityscapes).}
\label{tab:comp}
\end{table}

We also investigate whether CVRN performs stably with different panoptic segmentation architectures. We evaluated over two network architectures PFPN~\cite{kirillov2019PFPN} and UPSNet~\cite{xiong2019upsnet} that have been widely adopted in supervised panoptic segmentation. For each architecture, we trained 3 models including \textit{Source only}, \textit{MTST} and \textit{CVRN}, where \textit{MTST} is the base network that our proposed cross-view regularization is built upon. Table \ref{tab:ext} shown experimental results. It can be seen that \textit{CVRN} outperforms \textit{Source only} and \textit{MTST} with a large margin (over 4.6 in mPQ) in both panoptic segmentation architectures. Such experimental results show that our proposed cross-view regularization can work well with different panoptic segmentation architectures.

\renewcommand\arraystretch{1.2}
\begin{table}[t]
\centering
\begin{footnotesize}
\begin{tabular}{p{1.5cm}|*{2}{p{0.25cm}}p{0.45cm}|p{1.5cm}|*{2}{p{0.25cm}}p{0.45cm}}
\toprule
 \multicolumn{4}{c|}{\textbf{PFPN}} & \multicolumn{4}{c}{\textbf{UPSNet}} \\
 \midrule
Method &mSQ &mRQ &mPQ &Method &mSQ &mRQ &mPQ \\
\midrule
Source only  &54.3  &29.0  &21.5  &Source only  &58.7  &31.5  &23.3 \\
MTST  &64.8  &36.1  &26.8  &MTST  &66.2  &36.1  &27.5  \\
\textbf{CVRN} &\textbf{66.4} &\textbf{40.0} &\textbf{31.4}  &\textbf{CVRN} &\textbf{68.2}  &\textbf{43.4}  &\textbf{34.0}  \\
\bottomrule
\end{tabular}
\end{footnotesize}
\vspace{2pt}
\caption{CVRN works with different panoptic segmentation architectures well: \textit{CVRN} can work with different panoptic segmentation architectures (e.g. PFPN~\cite{kirillov2019PFPN} and UPSNet~\cite{xiong2019upsnet}) with consistent performance improvement as compared with \textit{Source only} and the baseline network \textit{MTST} (evaluated on domain adaptive panoptic segmentation task SYNTHIA $\rightarrow$ Cityscapes).}
\label{tab:ext}
\end{table}

\section{Conclusion}
This paper presents a cross-view regularization network that tackles domain adaptive panoptic segmentation by exploring an inter-task regularization (ITR) and an inter-style regularization (ISR). Specifically, ITR exploits the complementary nature of instance segmentation and semantic segmentation to regularize their self-training. ISR employs online image stylization to augments multiple views of the same image for self-training regularization. The proposed method have been evaluated extensively over multiple domain adaptive panoptic segmentation settings and experiments demonstrate its superior performance as compared with state-of-the-art. In the future, we will adapt the idea of cross-view regularization to adversarial learning, and explore cross-view structure regularization for better unsupervised domain adaptive panoptic segmentation.

\section*{Acknowledgement}
This research was conducted in collaboration with Singapore Telecommunications Limited and supported/partially supported (delete as appropriate) by the Singapore Government through the Industry Alignment Fund - Industry Collaboration Projects Grant.

{\small
\bibliographystyle{ieee_fullname}
\bibliography{egbib}
}

\end{document}